\documentclass[letterpaper, 10 pt, journal, twoside]{IEEEtran}  

\usepackage{amsmath}
\usepackage{amssymb}
\usepackage{xfrac}          

\usepackage{graphicx}

\usepackage{booktabs}       
\usepackage{multirow}       
\usepackage{hyperref}
\usepackage{url}
\title{\LARGE \bf
GAMMA: Gaussian Mixture Model Autoencoder for LiDAR Reconstruction

}

\author{Davide Possenti$^{1}$, Stefano Arrigoni$^{1}$, Jia Hu$^{2}$, Francesco Braghin $^{1}$
\thanks{$^{1}$D. Possenti, S. Arrigoni, and F. Braghin are with the Department of Mechanical Engineering, Politecnico di Milano, 20156 Milan, Italy 
        {\tt\small davide.possenti@polimi.it}}%
\thanks{$^{2}$J. Hu is with the Department of Transportation Engineering, Tongji University, Shanghai 201804, China.}%

\thanks{Supplementary video material at: \url{https://youtu.be/4iKNvUDkIdsB}}%
}

\begin{document}

\maketitle
\thispagestyle{empty}
\pagestyle{empty}


\begin{abstract}
Accurate 3D geometric understanding is essential for autonomous navigation, yet raw LiDAR scans are sparse, irregularly sampled, and sensor-dependent.
Furthermore, standard voxel-based encoders discard sub-voxel geometry and degrade under changes in beam density. We present GAMMA (Gaussian Mixture Model Autoencoder for LiDAR Reconstruction), a lightweight (1.4~M parameters) framework that encodes each scan into a latent volume and decodes continuous sub-voxel geometry in a single feed-forward pass at 20~FPS. A multi-scale GMM decoder predicts explicit mixture parameters, while occupancy-aware initialization and an exponential geometric-loss curriculum stabilize end-to-end training. Evaluated on the 64-beam View of Delft (VoD) dataset, GAMMA achieves F1@20\,cm\,=\,0.980 and L2 CD\,=\,0.0131\,$\mathrm{m}^2$. At the map level, pose-fused GAMMA predictions reach F1@20\,cm\,=\,94.93\%, compared with 87.39\% for SHINE-Mapping with a 20\,cm leaf resolution, while providing a $62.8\times$ speedup and reducing peak GPU memory and map storage by more than an order of magnitude. Finally, zero-shot evaluation confirms that the learned geometric prior transfers without fine-tuning across unseen 32-, 64-, and 128-beam LiDAR configurations, as well as to 4D radar (F1@20\,cm up to 0.871) despite an $86.9\times$ point-density gap.
\end{abstract}


\vspace{-4mm}

\section{Introduction}

\IEEEPARstart{A}{utonomous} vehicles and mobile robots rely on accurate
3D geometric understanding to navigate safely. LiDAR sensors are a standard
modality for this task~\cite{geiger2013vision}, yet raw point clouds pose
significant processing challenges: they are sparse, irregularly sampled, and
their density decreases rapidly with range. A useful geometric
representation should therefore preserve relevant scene structure while being
rapidly decodable and robust to variations in sensor sampling density.

Voxelization is commonly used to regularize this structure, enabling the use of
3D sparse convolutions~\cite{zhou2018voxelnet,yan2018second}. However,
this representation introduces two coupled limitations for geometric reconstruction.
First, discretization incurs \emph{quantization error}: geometric detail within
each voxel is collapsed to a single occupancy state, preventing sub-voxel recovery.
Second, voxel-based encoders learn features tightly coupled to sensor-specific
sampling patterns. Consequently, models trained on dense LiDAR configurations
(e.g., 64-beam setups~\cite{apalffy2022}) can degrade when deployed on sparser
sensors~\cite{xu2025mergeocc} (e.g., 32-beam~\cite{caesar2020nuscenes}) or on entirely different
sensing modalities such as 4D radar. This \emph{cross-sensor domain shift}
remains a major challenge in robotic perception. To recover continuous geometry, Implicit Neural Representations
(INRs)~\cite{mescheder2019occupancy,park2019deepsdf} offer high fidelity but
typically require dense spatial querying or scene-specific optimization.
Alternatively, Gaussian-based representations, from classical
Normal Distributions Transform (NDT)~\cite{biber2003normal} to recent neural splatting methods~\cite{kerbl20233d,hess2025splatad},
are a natural fit for LiDAR. Yet, modern neural Gaussian approaches typically
rely on per-scene gradient-based optimization, precluding the single-forward-pass
execution required by real-time robotic perception. We address these limitations with \textbf{GAMMA} (\textbf{GA}ussian \textbf{M}ixture
\textbf{M}odel \textbf{A}utoencoder for LiDAR Reconstruction). GAMMA uses a unified
\emph{sparse-to-dense} architecture: a sparse encoder maps each input scan into a
latent volume, while a multi-scale GMM decoder recovers continuous
sub-voxel geometry without expanding the spatial grid. GAMMA is trained end-to-end using occupancy supervision, Soft-F1 regularization and GMM negative
log-likelihood, predicting explicit Gaussian parameters (weights, means, covariances)
for fast analytical querying at inference. Unlike scene-specific neural mapping methods, GAMMA uses a fixed
pretrained model and requires no optimization when processing a new scan.  The main contributions are:
\begin{itemize}
    \item We propose a \emph{sparse-to-dense} autoencoder that maps
    heterogeneous point cloud inputs into a latent representation,
    reducing sensitivity to sensor-specific sampling patterns.

    \item We introduce a multi-scale GMM reconstruction head to recover
    continuous sub-voxel geometry, using an exponential activation schedule and
    occupancy-aware initialization to prevent component collapse.

    \item We validate GAMMA across baseline mapping and cross-sensor transfer.
    GAMMA outperforms SHINE-Mapping~\cite{zhong2023shine} in map accuracy ($94.9\%$ vs.\ $87.4\%$ F1@20\,cm)
    while requiring over an order of magnitude less runtime, GPU memory, and storage. Trained exclusively on
    64-beam LiDAR, the 1.4M-parameter model transfers zero-shot to unseen LiDAR configurations and to
    4D radar, without any fine-tuning.

\end{itemize}

\section{Related Work}
\subsection{LiDAR Scene Representation and 3D Sparse Convolutions}

Handling the irregular structure of large-scale outdoor point clouds has driven
two main paradigms. Point-based methods such as PointNet~\cite{qi2017pointnet}
and PointNet++~\cite{qi2017pointnet++} operate directly on raw points
without discretization, enabling permutation-invariant feature extraction.
However, their scalability to dense outdoor scans with hundreds of thousands of
points per frame is limited. Voxelization-based methods address this by
partitioning the scene into a regular grid and processing each occupied cell
with a compact feature encoder~\cite{zhou2018voxelnet}. The introduction of
3D sparse convolutions~\cite{yan2018second} further reduced the
computational cost by restricting operations to occupied voxels only, enabling
real-time 3D detection on full-resolution LiDAR frames. While these
architectures achieve strong supervised performance, their application to
\emph{self-supervised} scene reconstruction reveals a structural limitation:
the discrete voxel representation cannot recover the continuous geometry
within each cell, and the learned features are implicitly coupled to the
specific beam density of the training distribution. Our encoder adopts
the same efficient sparse backbone but projects its output onto a dense latent
volume, deliberately breaking the dependency on input sparsity patterns.

\subsection{3D Autoencoders and Latent Scene Representations}

Self-supervised learning of compact 3D representations has been explored
extensively. Early works such as FoldingNet~\cite{yang2018foldingnet} and
AtlasNet~\cite{groueix2018papier} decode latent codes into continuous 2D
manifolds folded in 3D space, achieving high-quality reconstruction of
single objects. More recent masked autoencoder approaches such as
Point-MAE~\cite{pang2023masked} and Point-BERT~\cite{yu2022point} show that
pre-training on point cloud patches yields powerful transferable
representations. To scale self-supervision to outdoor LiDAR, Voxel-MAE
~\cite{min2022voxel} and ALSO~\cite{boulch2023also} extend masked reconstruction
and occupancy-based learning to large-scale scenes. Similarly, semantic scene
completion methods such as SSCNet~\cite{song2017semantic} and LMSCNet~\cite{roldao2020lmscnet} infer dense volumetric
occupancy from sparse observations. However, these large-scale autoencoders
typically reconstruct fixed-resolution voxel occupancy or semantic occupancy,
which limits geometric fidelity due to quantization at voxel boundaries.
GAMMA bridges this gap by predicting multi-scale GMM parameters per voxel, enabling sub-voxel resolution in a single forward pass.

\subsection{Gaussian Reconstruction and Neural Implicit Mapping}

The use of Gaussian distributions to model 3D geometry has a long history in
point cloud registration and mapping. The Normal Distributions Transform
(NDT)~\cite{biber2003normal} represents local surface patches as Gaussians for
scan matching, while Eckart et~al.~\cite{eckart2016accelerated} extend GMMs to
efficient hierarchical point cloud compression. The recent emergence of 3D
Gaussian Splatting~(3DGS)~\cite{kerbl20233d} revived interest in Gaussian
representations for high-fidelity, real-time novel view synthesis, inspiring
extensions to autonomous driving: SplatAD~\cite{hess2025splatad} reconstructs
dynamic driving scenes for sensor simulation, while
LiDAR-GS~\cite{chen2024lidar} adapts the Gaussian paradigm to LiDAR simulation
with differentiable rendering. In parallel, scene-specific neural mapping methods such
as SHINE-Mapping~\cite{zhong2023shine} optimize an implicit representation
for each posed sequence. Despite their compelling quality, both neural Gaussian
and implicit mapping frameworks rely on per-scene iterative optimization,
requiring thousands of gradient steps to fit a single sequence. Unlike these
approaches, GAMMA predicts GMM parameters in a single forward pass via
a learned encoder-decoder.

\section{Methodology}

\begin{figure*}[t]
  \centering
  \includegraphics[width=0.9\textwidth]{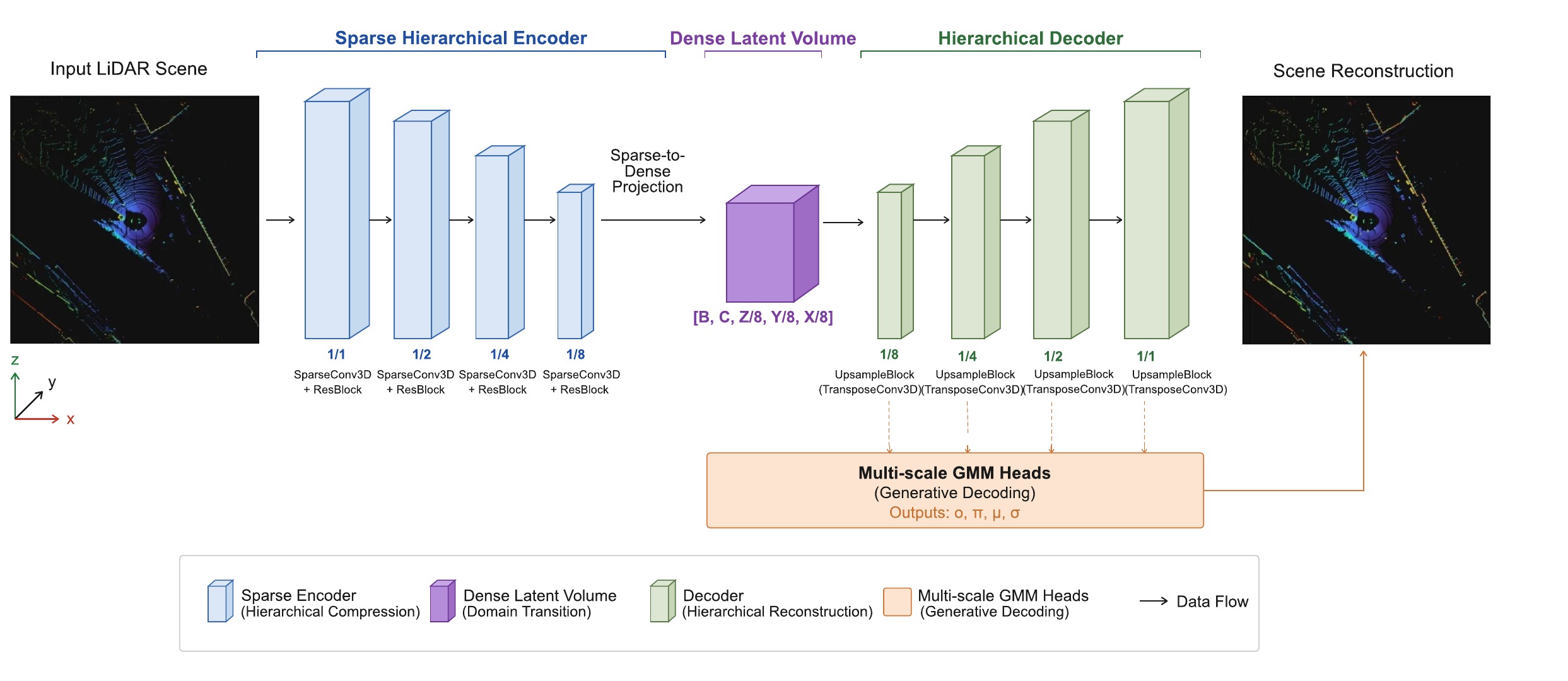} %
  \caption{GAMMA architecture overview. A sparse hierarchical encoder maps LiDAR inputs to a resolution-agnostic dense latent volume. A multi-scale hierarchical decoder predicts explicit GMM parameters ($o, \boldsymbol{\pi}, \boldsymbol{\mu}, \boldsymbol{\sigma}$) at multiple spatial scale to recover continuous sub-voxel geometry. }
  \label{fig:architecture}
\end{figure*}

As illustrated in Fig.~\ref{fig:architecture}, the proposed GAMMA framework consists of a sparse-to-dense encoder, a dense latent bottleneck, and a multi-scale GMM decoder for continuous sub-voxel reconstruction.

\subsection{Problem Formulation}

Let $\mathcal{P} = \{\mathbf{p}_i \in \mathbb{R}^3\}_{i=1}^{N}$ be a
raw LiDAR scan discretized into a voxel grid $\mathcal{V}$ of size
$X \!\times\! Y \!\times\! Z$, where each voxel spans a physical extent
$\boldsymbol{\delta} = [\delta_x, \delta_y, \delta_z]^\top$. GAMMA
learns $f_{\theta} : \mathcal{P} \mapsto \hat{\mathcal{P}}$ that encodes the input scan into a compact latent representation and
reconstructs a dense point cloud at sub-voxel resolution using only the
input scan as supervision.

\subsection{Sparse-to-Dense Encoder}

\subsubsection{Voxelization and Density Encoding}
Each occupied voxel $v$ is described by a 4-dimensional feature
\begin{equation}
    \mathbf{f}_v =
    \left[\bar{x}_v,\ \bar{y}_v,\ \bar{z}_v,\ \log(n_v)\right]
    \in \mathbb{R}^4,
\end{equation}
where $(\bar{x}_v, \bar{y}_v, \bar{z}_v)$ is the mean point coordinate
and $\log(n_v)$ is a logarithmic density descriptor, providing richer
information than binary occupancy alone.

\subsubsection{Sparse 3D Backbone}
The voxel features are lifted to a higher-dimensional embedding by a
per-voxel MLP composed of two Linear--BatchNorm--ReLU layers and then processed by a sparse
convolutional backbone with SubMConv3d and
strided SparseConv3d layers interleaved with residual blocks. The channel
progression is $\{16, 32, 64, C_z\}$ at spatial resolutions
$\{1, \sfrac{1}{2}, \sfrac{1}{4}, \sfrac{1}{8}\}$, with $C_z{=}64$.

\subsubsection{Sparse-to-Dense Projection}
At the final encoder stage, the sparse latent tensor is projected into a
dense volume
$\mathbf{Z} \in \mathbb{R}^{B \times C_z \times \frac{D}{8}
\times \frac{H}{8} \times \frac{W}{8}}$, separating efficient sparse
processing in the input domain from dense multi-scale decoding.

\subsection{Multi-Scale GMM Decoder with Sub-Voxel Precision}

\subsubsection{Decoder Architecture}

The dense latent representation is progressively upsampled by three
UpsampleBlock modules. Each block contains a transposed convolution with
kernel size 4 and stride 2, followed by a refinement convolution and a
3D residual block with GroupNorm and SiLU activations. The channel progression
is $C_z \rightarrow 32 \rightarrow 16 \rightarrow 8$, which restores the
original voxel-grid resolution. At each of the four decoder scales
$s \in \{\sfrac{1}{8}, \sfrac{1}{4}, \sfrac{1}{2}, 1\}$, a dedicated
$1 \times 1 \times 1$ GMM head predicts reconstruction parameters, providing
deep supervision throughout the decoding pathway.

\subsubsection{GMM Parameterization}

For each voxel at full resolution, the GMM head outputs $1 + 7K$ channels:
one occupancy logit, $K$ mixture-weight logits, $3K$ mean offsets, and
$3K$ dispersion parameters. The output is organized as
\begin{equation}
    \text{output} =
    \Bigl[o,\ \boldsymbol{\pi} \in \mathbb{R}^K,\ \boldsymbol{\mu} \in \mathbb{R}^{3K},\ \boldsymbol{\sigma} \in \mathbb{R}^{3K}\Bigr],
\end{equation}
where $o$ is the occupancy logit, $\boldsymbol{\pi}$ are mixture weights
normalized with a softmax, $\boldsymbol{\mu}$ are mean offsets expressed in
fractional voxel coordinates, and $\boldsymbol{\sigma}$ are per-axis standard
deviations parameterized with a softplus nonlinearity plus $\epsilon$ to ensure
positivity. The use of fractional voxel coordinates is intentional: the GMM models the
distribution of points inside each voxel in a local normalized frame rather
than in a global metric frame. This choice makes the decoder independent of
the absolute scene position and allows it to describe sub-voxel geometry
consistently across the full spatial extent.

At inference time, for each occupied voxel, the predicted Gaussian components
are converted back into metric coordinates according to
\begin{equation}
    \hat{\mathbf{p}}_{v,k} = \mathbf{c}_v + \boldsymbol{\mu}_{v,k} \odot \boldsymbol{\delta},
\end{equation}
where $\mathbf{c}_v$ is the center of voxel $v$ and
$\boldsymbol{\delta} = [\delta_x, \delta_y, \delta_z]^\top$ is the voxel-size
vector.

\subsection{Training Objective and Optimization Schedule}

\subsubsection{Multi-Scale Loss}

The total loss combines occupancy supervision and geometric supervision at
multiple decoder scales:
\begin{equation}
    \mathcal{L} =
    \sum_{s} w_s^{\text{occ}}
    \Bigl(\mathcal{L}^s_{\text{focal}} + \mathcal{L}^s_{\text{f1}}\Bigr)
    + \lambda_{\text{GMM}} \sum_{s} w_s^{\text{GMM}} \cdot \mathcal{L}^s_{\text{NLL}}.
\end{equation}
Here $\mathcal{L}_{\text{focal}}$ is the Sigmoid Focal Loss~\cite{lin2017focal} with
$\alpha = 0.4$ and $\gamma = 1.0$, used to address the severe class imbalance
of outdoor LiDAR occupancy prediction. $\mathcal{L}_{\text{f1}}$ is a soft
Dice-style loss~\cite{milletari2016v} that directly improves the overlap between predicted and
target occupancy. $\mathcal{L}_{\text{NLL}}$ is the Gaussian Mixture
Negative Log-Likelihood computed only at ground-truth point locations within
occupied voxels. Crucially, the NLL is evaluated in the local normalized voxel frame. Let
$\mathbf{c}_{v(i)}$ be the center of the voxel containing point
$\mathbf{p}_i$, and let $\boldsymbol{\delta}$ be the voxel-size vector.
The point is mapped to fractional voxel coordinates via
\begin{equation}
    \tilde{\mathbf{p}}_i =
    \frac{\mathbf{p}_i - \mathbf{c}_{v(i)}}{\boldsymbol{\delta}},
\end{equation}
where the division is element-wise. The GMM parameters are predicted in the
same normalized coordinate system, so the likelihood can be written as
\begin{equation}
    \mathcal{L}_{\text{NLL}} =
    -\frac{1}{N} \sum_{i=1}^{N}
    \log \sum_{k=1}^{K} \pi_k \cdot
    \mathcal{N}\!\left(
        \tilde{\mathbf{p}}_i \;\middle|\;
        \boldsymbol{\mu}_k,\ \mathrm{diag}(\boldsymbol{\sigma}_k^2)
    \right).
\end{equation}
This formulation models the intra-voxel geometry in a scale-normalized space
while preserving a metric interpretation through the final reconstruction
step.

The scale weights are set to
$w_s^{\text{occ}} \in \{1.0, 0.5, 0.25, 0.125\}$ and
$w_s^{\text{GMM}} \in \{0.9, 0.05, 0.03, 0.02\}$ for
$s \in \{1,2,4,8\}$, which concentrates geometric supervision at the finest
resolution while retaining multi-scale regularization.

\subsubsection{Exponential GMM Ramp-Up Schedule}
For the first $e_0$ epochs, the GMM weight is
set to zero and the model focuses exclusively on learning occupancy structure. Afterwards it increases as:
\begin{equation}
    \lambda_{\text{GMM}}(e) =
    \lambda_{\max}\Bigl(1 - e^{-5t}\Bigr),
    \quad
    t = \frac{e - e_0}{e_{\mathrm{warm}}},
\end{equation}
with $\lambda_{\max}{=}0.3$ and $e_{\mathrm{warm}}{=}30$ epochs. This
curriculum prevents the GMM objective from dominating optimization before
the occupancy head has converged.

\subsubsection{Occupancy Prior Bias Initialization}
To improve early convergence, the occupancy output bias is initialized to:
\begin{equation}
    b_0 = -\log\!\left(\frac{1 - p_{\text{prior}}}{p_{\text{prior}}}\right),
\end{equation}
where $p_{\text{prior}}$ is the empirical voxel occupancy rate. GMM variance biases are initialized to a small positive
value to ensure stable initial covariances.

\section{Experimental Evaluation}
\label{sec:experiments}

\subsection{Experimental Setup}

\subsubsection{Datasets}
We train and validate GAMMA exclusively on the \textbf{View of Delft (VoD)} dataset~\cite{apalffy2022}, which provides synchronized frames from a 64-beam Velodyne HDL-64E LiDAR. We use a dataset split of 6{,}647 training, 739 validation, and 1{,}296 test frames. All point clouds are clipped to the egocentric extent $[-51.2,51.2]\!\times\![-51.2,51.2]\!\times\![-2,2]$~m and discretized on a $10\!\times\!10\!\times\!50$~cm voxel grid. The chosen spatial extent is relatively large for an autoencoding setting, but it is appropriate for autonomous driving, where long-range context, road layout, and distant agents are important for faithful scene reconstruction. To evaluate zero-shot generalization across sensors and domains, we directly test the model on four datasets completely unseen during training, validation, checkpoint selection, and hyperparameter tuning, without any fine-tuning or domain adaptation: \textbf{KITTI}~\cite{geiger2013vision} (64-beam, 1{,}059 frames), \textbf{Waymo Open Dataset}~\cite{sun2020scalability} (64-beam, 191 frames), \textbf{nuScenes}~\cite{caesar2020nuscenes} (32-beam, 3{,}376 frames), and \textbf{MSC-RAD4R}~\cite{choi2023msc} (128-beam, 3{,}453 frames).  We additionally evaluate cross-modality transfer
to 4D radar point clouds from VoD and MSC-RAD4R.

\subsubsection{Evaluation Protocol}
GAMMA is evaluated per frame: each scan is
independently voxelized, encoded, decoded, and compared with its input, and metrics are averaged over the evaluation split. GAMMA uses no recurrent state or neighboring observations. Moreover, for the SHINE-Mapping comparison, independently decoded GAMMA scans are pose-aligned and fused into a global map, whereas SHINE optimizes a scene-specific map from the posed sequence. For multi-scan evaluation, consecutive pose-aligned LiDAR scans are aggregated and processed as one independent input.

\subsubsection{Evaluation Metrics}
We assess reconstruction quality using complementary geometric and distributional metrics.
First, \textbf{Point-cloud F1}, precision, and recall are computed at strict distance thresholds $\tau_d \in \{10, 15, 20\}$~cm to quantify local structural accuracy.
Second, we report the \textbf{Squared Chamfer Distance (L2 CD)}, measured in $\mathrm{m}^2$, which penalizes larger spatial deviations more strongly than the $L_1$ variant and is therefore well suited to LiDAR reconstruction under localization errors. This metric is also consistent with our Gaussian Mixture negative log-likelihood objective, which explicitly models spatial spread.
Third, we compute \textbf{Jensen-Shannon Divergence (JSD)} and \textbf{Sliced Wasserstein Distance (SWD)} to assess distributional similarity and density preservation.
Finally, \textbf{Voxel IoU} is used to measure coarse binary occupancy.
Inference throughput is reported in frames per second (FPS) on a single NVIDIA RTX~5090 GPU.

\subsubsection{Implementation Details}
The default GAMMA configuration uses a latent dimension $C_z{=}64$, $K{=}3$ GMM components, a ramp-up start epoch $e_0{=}12$. The network is trained for 100 epochs using AdamW~\cite{loshchilov2017decoupled} (learning rate $10^{-4}$, weight decay $10^{-4}$) and a cosine annealing scheduler~\cite{loshchilov2016sgdr}. The total parameter count is 1.4~M. We select the final checkpoint based on the best validation F1@0.45.

\subsection{Main Results on the Target Domain}

\begin{figure*}[t]
  \centering
  \includegraphics[width=\textwidth]{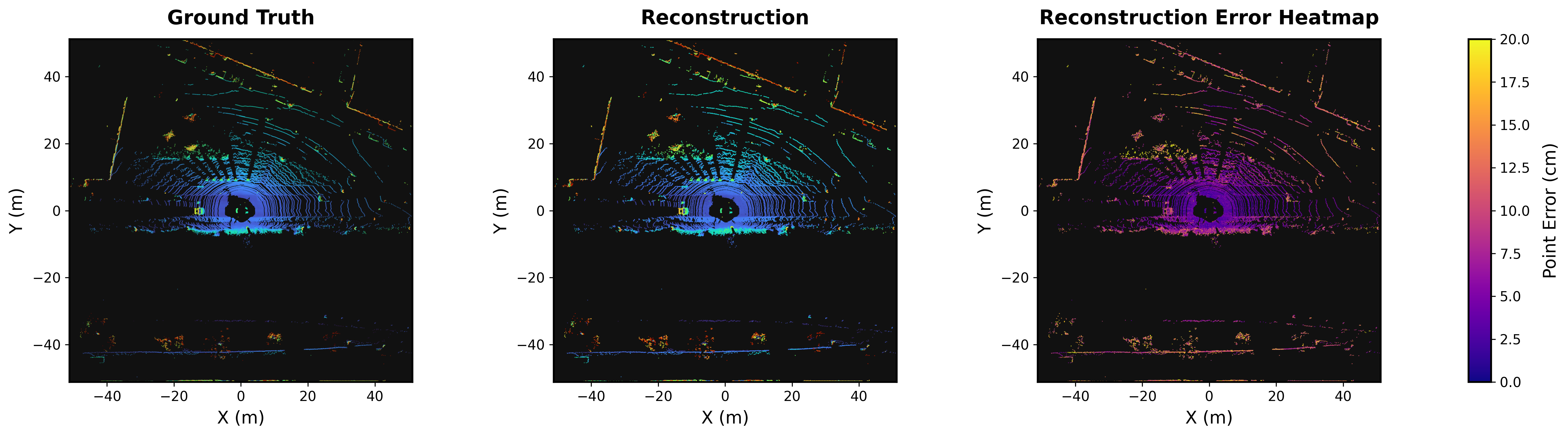}
  \caption{Qualitative reconstruction on a View of Delft (VoD) test frame. \textbf{Left and Center:} Ground truth and GAMMA reconstruction. Points are colored by elevation ($Z \in [-2, 2]$~m) to highlight physical scene structures such as vehicles and infrastructure. \textbf{Right:} Point-to-point $L_2$ error heatmap. }
  \label{fig:qualitative_vod}
\end{figure*}

Table~\ref{tab:main_vod} reports reconstruction performance on the VoD
test set (1{,}296 frames, $\tau_{\rm occ}{=}0.50$).
GAMMA achieves strong geometric fidelity across all thresholds
(F1@20\,cm\,=\,0.980, F1@15\,cm\,=\,0.943, F1@10\,cm\,=\,0.823) with precision and recall of 0.970 and 0.990 at the 20cm tolerance.
At the strictest threshold, recall (0.872) outpaces precision (0.781),
consistent with the sub-voxel uncertainty modelled by the GMM.
A qualitative visualization is provided in
Fig.~\ref{fig:qualitative_vod}. 

Table~\ref{tab:fulltest} provides a complementary statistical
characterization.
The mean L2 CD is 0.0131~$\mathrm{m}^2$ with an average point ratio
of 0.862, confirming that the reconstructed clouds slightly
underestimate ground-truth density without systematic over-generation.
Metric variance is low across all thresholds (F1@20\,cm: $0.9798 \pm 0.0053$,\
 F1@15\,cm: $0.9433 \pm 0.013$,\
 F1@10\,cm: $0.8233 \pm 0.035$), demonstrating consistent
geometric fidelity across diverse scenes.
The JSD of 0.2243 and SWD of 4.8798 confirm that GAMMA preserves
the global density structure and statistical coherence of the scan.

\begin{table}[t]
  \caption{GAMMA reconstruction performance on the VoD test set,
    $\tau_{\rm occ}{=}0.50$.}
  \label{tab:main_vod}
  \centering
  \begin{tabular}{lccc}
    \toprule
    $\tau_d$ & F1 Score & Precision & Recall \\
    \midrule
    10~cm & 0.823 & 0.781 & 0.872  \\
    15~cm & 0.943 & 0.917 & 0.972  \\
    20~cm & \textbf{0.980} & \textbf{0.970} & \textbf{0.990} \\
    \bottomrule
  \end{tabular}
\end{table}

\begin{table}[t]
  \caption{Full-test statistics on VoD test set (1{,}296 frames).}
  \label{tab:fulltest}
  \centering
  \setlength{\tabcolsep}{4pt}
  \begin{tabular}{lrrrr}
    \toprule
    Metric & Mean & Std & Min & Max \\
    \midrule
    L2 CD ($\mathrm{m}^2$) & 0.0131 & 0.0024 & 0.0077 & 0.0397 \\
    Ratio                  & 0.8620 & 0.1538 & 0.4233 & 1.2071 \\
    JSD                    & 0.2243 & 0.0380 & 0.1625 & 0.3555 \\
    SWD                    & 4.8798 & 0.8352 & 3.2758 & 7.1212 \\
    F1@10cm                & 0.8233 & 0.0350 & 0.7273 & 0.9138 \\
    F1@15cm                & 0.9433 & 0.0128 & 0.9176 & 0.9774 \\
    F1@20cm                & 0.9798 & 0.0053 & 0.9667 & 0.9922 \\
    \bottomrule
  \end{tabular}
\end{table}

\subsection{Distance-Dependent Reconstruction Quality}
\label{ssec:range}

Table~\ref{tab:range} decomposes the VoD test-set performance into three
radial bands relative to the ego-vehicle. Near-field reconstruction
(0--15~m) is excellent: F1@10cm reaches 0.900 and F1@20cm 0.994, with a
mean L2 CD of only 0.0082~$\mathrm{m}^2$.
Performance degrades progressively with distance, as expected in sparse
LiDAR scans, but GAMMA continues to recover the scene geometry robustly.
In the mid range (15--30~m), F1@10cm drops to 0.763 while F1@20cm remains
high at 0.974, indicating that the model still captures the coarse scene
structure very well. In the far field ($>$30~m), F1@10cm falls to 0.555,
yet F1@20cm remains at 0.910, showing that the global layout of the scene
is still largely preserved. The monotonic increase in L2 CD across the
three bands $(0.0082, 0.0164, 0.0351~\mathrm{m}^2)$ confirms a gradual
and predictable degradation rather than a sharp failure mode. This behavior reflects the physical divergence of LiDAR beams at long range:
distant regions are inherently geometrically undersampled, making centimeter-level surface estimation fundamentally ill-posed. Crucially,
global scene structure remains highly reliable at decimeter tolerance ($\mathrm{F1@20\,cm}=0.910$ beyond 30\,m), providing an effective geometric
prior for coarse occupancy reasoning, path planning, and compact map transmission without requiring dense point-level recovery.

\begin{table}[t]
  \caption{Range-dependent reconstruction quality on VoD test set
    ($\tau_{\rm occ}{=}0.50$, 1{,}296 frames).}
  \label{tab:range}
  \centering
  \begin{tabular}{lccc}
    \toprule
    Range & F1@10cm & F1@20cm & L2 CD ($\mathrm{m}^2$) \\
    \midrule
    Near (0--15~m) & \textbf{0.900} & \textbf{0.994} & \textbf{0.0082} \\
    Mid (15--30~m) & 0.763          & 0.974          & 0.0164 \\
    Far ($>$30~m)  & 0.555          & 0.910          & 0.0351 \\
    \bottomrule
  \end{tabular}
\end{table}


\subsection{Multi-Scan Aggregation and Pose Sensitivity}
\label{sssec:multiscan}

\begin{figure*}[t]
  \centering
  \includegraphics[width=0.85\textwidth]{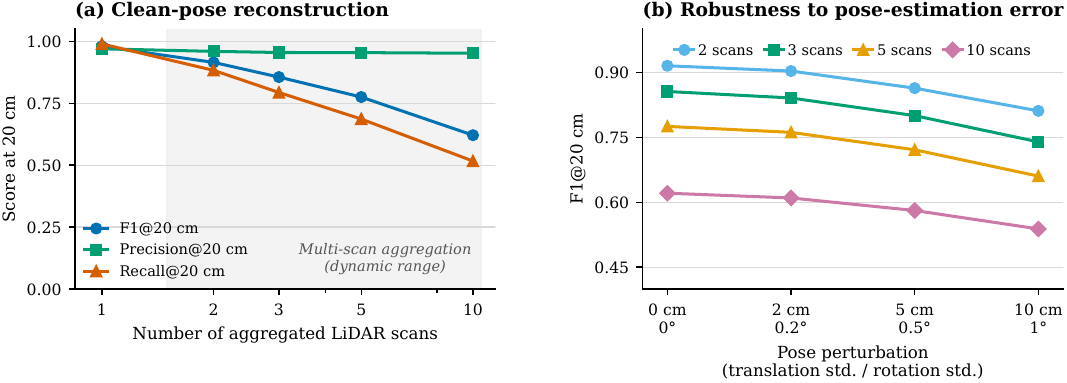}
  \caption{Multi-scan LiDAR aggregation and pose sensitivity.
  \textbf{(a)} Clean-pose reconstruction quality as the number of
  pose-aligned scans increases. The shaded region denotes multi-scan inputs
  with dynamically expanded spatial extent.
  \textbf{(b)} F1@20\,cm under synthetic pose perturbations. Each level gives
  the standard deviation of translation and rotation noise applied to the
  relative poses used for scan alignment.}
  \label{fig:multiscan}
\end{figure*}

We evaluate GAMMA on inputs formed by aggregating
$N\in\{2,3,5,10\}$ consecutive VoD LiDAR scans into a common ego frame
using the available poses. Each aggregate is processed as a single
independent input without architectural changes or retraining. The average number of input points grows from 79{,}852 for a single scan to 798{,}323 for $N{=}10$. Thus, the $N{=}10$ condition requires reconstructing nearly 0.8~million points in a single forward pass. Figure~\ref{fig:multiscan}(a) shows that, under clean pose alignment,
F1@20\,cm decreases from 0.980 for one scan to 0.621 for ten scans.
Precision remains high ($\geq0.952$), whereas recall decreases from 0.990
to 0.516. The loss is therefore dominated by reduced coverage of the
substantially denser, dynamically expanded input rather than by spurious
geometric generation. To assess pose sensitivity, we perturb the relative alignment poses with
zero-mean translation and rotation noise. Figure~\ref{fig:multiscan}(b)
reports F1@20\,cm for perturbation standard deviations of
$(0\,\mathrm{cm},0^\circ)$, $(2\,\mathrm{cm},0.2^\circ)$,
$(5\,\mathrm{cm},0.5^\circ)$, and $(10\,\mathrm{cm},1^\circ)$.
At the largest perturbation, F1@20\,cm is 0.811, 0.739, 0.660, and 0.538
for $N{=}2$, $3$, $5$, and $10$, respectively. These results show that
pose uncertainty increasingly affects longer aggregation windows, whereas
short windows remain comparatively robust without temporal training or
recurrent state.

\subsection{Ablation Studies}
\label{ssec:ablation}

\subsubsection{Component Ablation}
Table~\ref{tab:ablation_comp} reports the contribution of each architectural choice.

\noindent\textbf{GMM reconstruction objective.}
Removing the GMM branch causes a large drop in reconstruction quality: F1@0.45 decreases from 0.9785 to 0.8371 and L2 CD worsens from 0.0118 to 0.0402~$\mathrm{m}^2$. This confirms that occupancy supervision alone is insufficient to recover fine geometry. Notably, IoU improves (0.9388 vs.\ 0.8813), because the occupancy-only model learns to over-predict voxel occupancy
aggressively to maximize the binary overlap signal, at the cost of
geometric precision. 

\noindent\textbf{Point density encoding.}
The log-density feature $\log(n_v)$ provides local return-density information to the encoder. Removing it slightly reduces F1@0.45 from 0.9785 to 0.9691, while leaving L2 CD almost unchanged. This indicates that density encoding mainly helps sub-voxel precision in regions with strong sparsity variation.

\noindent\textbf{Number of mixture components.}
Using a single Gaussian ($K{=}1$) already restores most of the lost fidelity (F1@0.45 = 0.9703), but $K{=}3$ yields the best overall trade-off, improving both F1 and L2 CD. This suggests that multiple components are beneficial for modeling the multimodal structure that appears at boundaries and object edges.

\begin{table}[t]
  \caption{Component ablation on VoD validation set (best epoch). $\bigstar$ denotes the default configuration.}
  \label{tab:ablation_comp}
  \centering
  \setlength{\tabcolsep}{4pt}
  \begin{tabular}{lcccc}
    \toprule
    Model Config & Best Ep. & F1@0.45 & L2 CD ($\mathrm{m}^2$) & IoU \\
    \midrule
    No GMM (occ. only)        & 90 & 0.8371 & 0.0402 & 0.9388 \\
    GAMMA, no density feat.   & 88 & 0.9691 & 0.0131 & 0.8679 \\
    GAMMA, $K{=}1$            & 93 & 0.9703 & 0.0133 & 0.8767 \\
    \textbf{GAMMA, $K{=}3$} $\bigstar$ & 99 & \textbf{0.9785} & \textbf{0.0118} & \textbf{0.8813} \\
    \bottomrule
  \end{tabular}
  \smallskip
\end{table}

\subsubsection{GMM Activation Epoch ($e_0$)}
In our training schedule, the geometric objective is activated after
$e_0$ epochs and ramped up exponentially over $e_{\rm warm}{=}30$
epochs with maximum weight 0.3. Sweeping $e_0 \in \{10, 11, 12, 13,
14\}$ shows negligible variation across all metrics, confirming that
the procedure is robust to this hyperparameter. Among the tested
values, $e_0{=}12$ achieves the best joint optimum, with the highest
F1@0.45 and IoU alongside the lowest L2 CD, and is therefore adopted as
the default configuration.

\subsubsection{Voxel Resolution}
Table~\ref{tab:ablation_voxel} compares three voxel configurations on the VoD test set with $e_0{=}12$ fixed. The XY cell size is the primary driver of reconstruction quality: as the grid becomes coarser, accuracy decreases while throughput improves, showing a clear trade-off between fidelity and speed. The default $10\!\times\!10\!\times\!50$~cm grid offers the best balance, achieving the highest F1@10cm and lowest L2 CD at real-time speed. The intermediate $20\!\times\!20\!\times\!25$~cm setting is a useful compromise when strict 10~cm precision is less critical, while the coarsest grid is suitable for latency-sensitive use cases.

\begin{table}[t]
  \caption{Voxel resolution ablation on VoD test set ($\tau_{\rm occ}{=}0.50$, $C_z{=}64$, $e_0{=}12$). $\bigstar$ denotes the default configuration.}
  \label{tab:ablation_voxel}
  \centering
  \begin{tabular}{lccccc}
    \toprule
    Config (cm) & F1@10 & F1@15 & F1@20 & L2 CD ($\mathrm{m}^2$) & FPS \\
    \midrule
    $10\!\times\!10\!\times\!50$ $\bigstar$ & \textbf{0.823} & 0.943 & 0.980 & \textbf{0.0131} & 20 \\
    $20\!\times\!20\!\times\!25$            & 0.7638 & \textbf{0.9593} & \textbf{0.9925} & 0.0142 & 33 \\
    $20\!\times\!20\!\times\!50$            & 0.5967 & 0.8065 & 0.9271 & 0.0272 & \textbf{67} \\
    \bottomrule
  \end{tabular}
\end{table}


\subsection{Map-Level Benchmark against Neural Mapping}
\label{ssec:shine}

\begin{table*}[t]
  \caption{Map-level comparison on four VoD test sequences (1{,}296 frames). The 10\,cm and
  20\,cm labels in SHINE rows denote its leaf resolution. Peak GPU memory and
  persistent storage use binary units (GiB and MiB).}
  \label{tab:shine}
  \centering
  \scriptsize
  \setlength{\tabcolsep}{2.6pt}
  \resizebox{\textwidth}{!}{%
  \begin{tabular}{lcccccccccccc}
    \toprule
    Method
    & MAE Acc. (m) $\downarrow$
    & MAE Comp. (m) $\downarrow$
    & CD-L1 (m) $\downarrow$
    & Precision $\uparrow$
    & Recall $\uparrow$
    & F1@10\,cm $\uparrow$
    & F1@15\,cm $\uparrow$
    & F1@20\,cm $\uparrow$
    & Runtime (s) $\downarrow$
    & Peak GPU (GiB) $\downarrow$
    & Storage (MiB) $\downarrow$ \\
    \midrule
    \textbf{GAMMA}
    & \textbf{0.0398}
    & \textbf{0.1395}
    & \textbf{0.0896}
    & \textbf{99.58\%}
    & \textbf{90.94\%}
    & \textbf{89.94\%}
    & \textbf{93.93\%}
    & \textbf{94.93\%}
    & \textbf{81.87}
    & \textbf{2.23}
    & \textbf{16.57} \\
    SHINE-Mapping (20\,cm)
    & 0.0796
    & 0.1970
    & 0.1383
    & 94.13\%
    & 82.31\%
    & 67.78\%
    & 80.73\%
    & 87.39\%
    & 5139.07
    & 31.01
    & 1051.94 \\
    SHINE-Mapping (10\,cm)
    & 0.0494
    & 0.2307
    & 0.1400
    & 99.29\%
    & 79.85\%
    & 74.38\%
    & 84.04\%
    & 88.19\%
    & 9334.26
    & 31.34
    & 5053.21 \\
    \bottomrule
  \end{tabular}%
  }
\end{table*}

We compare GAMMA with SHINE-Mapping~\cite{zhong2023shine}, a scene-specific
implicit neural mapping baseline. This benchmark differs from the per-frame
autoencoding experiments: GAMMA independently decodes each scan with a fixed
pretrained model, transforms the decoded points into a common global frame
using the available poses, and fuses them per sequence, whereas SHINE-Mapping
optimizes one implicit representation for each posed sequence. Both methods are evaluated on the same four VoD test sequences (delft\_01,
delft\_10, delft\_14, and delft\_22), together comprising the same
1{,}296 test frames used elsewhere in this paper, against the same fused
raw LiDAR reference maps. Under a shared map-level protocol, predictions are
cropped to the reference bounding volume and both maps are voxelized at
2\,cm; prediction-to-reference distances above 0.40\,m are discarded,
whereas reference-to-prediction distances are truncated at 2.0\,m.

Table~\ref{tab:shine} shows that GAMMA outperforms both SHINE configurations
across all reported geometric metrics. At 20\,cm, GAMMA reaches
F1\,=\,94.93\%, compared with 87.39\% for SHINE-Mapping with a 20\,cm leaf
resolution and 88.19\% with a 10\,cm leaf resolution. GAMMA also achieves the
lowest MAE accuracy (0.0398\,m), MAE completeness (0.1395\,m), and CD-L1
(0.0896\,m), while attaining the highest precision and recall. The same
ordering is retained at the stricter 10\,cm and 15\,cm thresholds. The advantage over
SHINE-20\,cm is consistent for every sequence: GAMMA attains 96.83\% versus
89.12\% on delft\_01 (+7.71 percentage points), 98.51\% versus 92.85\% on
delft\_10 (+5.67), 95.93\% versus 89.02\% on delft\_14 (+6.90), and 88.44\%
versus 78.57\% on delft\_22 (+9.88). The feed-forward formulation also reduces map-construction cost. Runtime,
measured over all four sequences and excluding metric computation, is
81.87\,s for GAMMA versus 5{,}139.07\,s for SHINE-20\,cm and 9{,}334.26\,s
for SHINE-10\,cm, corresponding to approximately $63\times$ and
$114\times$ longer runtimes, respectively. GAMMA uses 2.23\,GiB peak
GPU memory and 16.57\,MiB persistent storage, compared with
31.01\,GiB/1{,}051.94\,MiB for SHINE-20\,cm and
31.34\,GiB/5{,}053.21\,MiB for SHINE-10\,cm. These results show that GAMMA achieves higher map-level reconstruction fidelity while avoiding
the per-sequence optimization cost of implicit neural mapping.

\subsection{Zero-Shot Generalization across LiDAR and 4D Radar}
\label{ssec:zeroshot}

\subsubsection{Cross-LiDAR Transfer}

Table~\ref{tab:zeroshot} evaluates zero-shot generalization under
simultaneous shifts in geography, scene topology, sensor mounting,
and beam density. On KITTI, performance slightly exceeds
the VoD target-domain operating point, confirming that the learned
representation transfers directly across compatible 64-beam setups.
Waymo remains robust, with only a moderate decrease at the strict 10\,cm
threshold. Remarkably, MSC-RAD4R achieves strong results despite
the 128-beam acquisition geometry differing substantially from the
training sensor (F1@20\,cm\,=\,0.931, F1@10\,cm\,=\,0.744).
The most challenging transfer is
nuScenes, where the 32-beam configuration produces a sparser reference cloud and a
higher predicted-to-GT density ratio (1.11); nevertheless, F1@20\,cm reaches
0.873, confirming that global surface geometry is preserved across
diverse sensor configurations without fine-tuning.

\begin{table}[t]
  \caption{Zero-shot reconstruction results across diverse datasets ($\tau_{\rm occ}{=}0.50$). Ratio indicates $|\hat{\mathcal{P}}|/|\mathcal{P}|$.}
  \label{tab:zeroshot}
  \centering
  \resizebox{\columnwidth}{!}{%
  \begin{tabular}{lcccccc}
    \toprule
    Dataset & Beams & F1@10cm & F1@15cm & F1@20cm & L2 CD ($\mathrm{m}^2$) & Ratio \\
    \midrule
    VoD (Target) & 64  & 0.823 & 0.943 & 0.980 & 0.0131 & 0.86 \\
    KITTI        & 64  & 0.825 & 0.948 & 0.982 & 0.0128 & 0.66 \\
    Waymo        & 64  & 0.748 & 0.908 & 0.962 & 0.0190 & 0.97 \\
    nuScenes     & 32  & 0.703 & 0.812 & 0.873 & 0.0353 & 1.11 \\
    MSC-RAD4R    & 128 & 0.744 & 0.882 & 0.931 & 0.0248 & 0.79 \\
    \bottomrule
  \end{tabular}%
  }
\end{table}

\subsubsection{Cross-Modality Transfer to 4D Radar}

We further evaluate zero-shot transfer from LiDAR to 4D radar, without any
radar fine-tuning. Due to the substantially lower radar density, we report
F1@20\,cm as a decimeter-level geometric consistency metric. For VoD, five
consecutive radar scans are ego-aligned and aggregated; for MSC-RAD4R, radar
scans are processed independently. On VoD frames, radar contains
918.8 points on average versus 79{,}852 LiDAR points ($86.9\times$ sparser).
Despite this gap, GAMMA achieves an F1@20\,cm of 0.871 on VoD radar and
0.800 on MSC-RAD4R (Table~\ref{tab:radar_zeroshot}). In both cases, recall
exceeds precision (0.915 vs.\ 0.832 on VoD, 0.826 vs.\ 0.778 on MSC-RAD4R), alongside predicted-to-target point ratios of
3.26 and 2.62, respectively. This behavior is consistent with decoding a
denser, continuous geometric surface from sparse radar returns rather than
generating spurious artifacts.

\begin{table}[t]
  \caption{Zero-shot transfer to 4D radar at 20\,cm tolerance
  ($\tau_{\rm occ}{=}0.50$).}
  \label{tab:radar_zeroshot}
  \centering
  \resizebox{\columnwidth}{!}{%
  \begin{tabular}{lccccc}
    \toprule
    Dataset  & F1@20\,cm & Prec.@20 & Rec.@20 & L2 CD ($\mathrm{m}^2$) & Ratio \\
    \midrule
    VoD Radar
     & 0.871 & 0.832 & 0.915 & 0.0919 & 3.26 \\
    MSC-RAD4R Radar
     & 0.800 & 0.778 & 0.826 & 0.1052 & 2.62 \\
    \bottomrule
  \end{tabular}%
  }
\end{table}

\section{Conclusions}

We presented GAMMA, a lightweight (1.4~M parameters) sparse-to-dense autoencoder
for feed-forward 3D scene reconstruction at 20~FPS. By decoding continuous
sub-voxel geometry via a multi-scale Gaussian mixture model, GAMMA achieves
F1@20\,cm\,=\,0.980 and L2 $\text{CD}$\,=\,0.0131\,m$^2$ on View of Delft. Compared to SHINE-Mapping, GAMMA attains higher map accuracy
($94.9\%$ vs.\ $87.4\%$ F1) while reducing reconstruction time (81.9\,s vs.\
5139.1\,s), peak GPU memory (2.23 vs.\ 31.01\,GiB), and storage (16.6 vs.\
1051.9\,MiB). The learned geometric prior also
generalizes zero-shot to different LiDAR configurations and to 4D radar,
without any fine-tuning. Future work will address sensor-dependent output density through conditioned decoding, enhance fine vertical structure in sparse far-range regions using range-adaptive anisotropic voxelization with range-aware loss reweighting, and recover point-level fidelity under substantial domain shifts via targeted fine-tuning.

\section*{APPENDIX}
\section{Storage Footprint and Latent Precision}
\label{app:memory}

We evaluate storage footprint and latent precision on all
1{,}296 VoD test frames. All reconstruction results in the main
paper use FP32 latent features; FP16 and INT8 storage are evaluated
here with the same pretrained model, without retraining.
Table~\ref{tab:storage_appendix} reports the average per-frame
footprint. Raw XYZ payloads contain three coordinates per point.
GAMMA payloads include the feature vectors and three INT16 spatial
indices per active latent cell, counted for all storage formats.
INT8 also includes one FP32 scale factor per feature channel,
required to recover floating-point features before decoding.

Direct FP16 coordinate storage provides a 2$\times$ reduction
relative to raw FP32 XYZ. Against this baseline, GAMMA FP16
requires 406.10 rather than 467.88~KiB/frame, a further 13.2\%
reduction, with the reconstruction error associated with its
learned representation.
GAMMA INT8 increases the storage saving relative to raw XYZ
FP16 to 54.6\%, requiring 212.39~KiB/frame. To isolate the additional effect of reduced-precision storage,
Table~\ref{tab:quant_appendix} compares FP32, FP16, and INT8
versions of the same latent features. Features are cast to FP16
or symmetrically quantized per channel to INT8, then converted
back to FP32 before decoding; spatial indices remain unchanged.
FP16 leaves all reported metrics unchanged.
Notably, INT8 also closely preserves FP32 reconstruction accuracy
without retraining: the displayed F1 scores decrease by only
0.0001 at 10 and 15\,cm, while F1@20\,cm and L2 CD remain
unchanged at the same precision.
We therefore recommend INT8 when storing GAMMA latent features:
the complete per-frame payload decreases from 794.01 to
212.39~KiB while reconstruction quality remains nearly identical
to FP32.

Finally, Table~\ref{tab:storage_appendix} also includes binary
OctoMap~\cite{hornung2013octomap} as an occupancy-based storage
reference. Since its isotropic cells cannot directly match
GAMMA's anisotropic $10 \times 10 \times 50$\,cm discretization,
we report several leaf sizes rather than a single matched setting.
At 10\,cm leaf resolution, OctoMap occupies 1{,}192.22~MiB, $4.43\times$ the GAMMA INT8 footprint. At a coarser 15\,cm leaf size, OctoMap requires
407.85~MiB, whereas GAMMA INT8 remains 34.1\% more compact. Among the evaluated settings, OctoMap achieves a smaller footprint
only at a 20\,cm leaf size (190.64~MiB). However, it stores only
binary cell occupancy, whereas GAMMA
also reconstructs continuous sub-voxel geometry.

\begin{table}[t]
  \caption{Average storage footprint on VoD.
  Ratio is raw FP32 XYZ size divided by representation size.}
  \label{tab:storage_appendix}
  \centering
  \setlength{\tabcolsep}{6pt}
  \small
  \begin{tabular}{lrr}
    \toprule
    Representation & KiB/frame & Ratio \\
    \midrule
    Raw XYZ FP32 & 935.77 & 1.00$\times$ \\
    Raw XYZ FP16 & 467.88 & 2.00$\times$ \\
    \midrule
    GAMMA FP32 & 794.01 & 1.18$\times$ \\
    GAMMA FP16 & 406.10 & 2.30$\times$ \\
    GAMMA INT8 & 212.39 & 4.41$\times$ \\
    \midrule
    OctoMap (10\,cm)   & 942.0 & 0.99$\times$ \\
    OctoMap (12.5\,cm) & 520.7 & 1.80$\times$ \\
    OctoMap (15\,cm)   & 322.9 & 2.90$\times$ \\
    OctoMap (20\,cm)   & 151.1 & 6.21$\times$ \\
    \bottomrule
  \end{tabular}
\end{table}

\begin{table}[t]
  \caption{GAMMA latent storage precision ablation on VoD.}
  \label{tab:quant_appendix}
  \centering
  \setlength{\tabcolsep}{3pt}
  \small
  \begin{tabular}{lcccc}
    \toprule
    Format & F1@10 & F1@15 & F1@20 & L2 CD \\
    \midrule
    GAMMA FP32 & 0.8233 & 0.9433 & 0.9798 & 0.0131 \\
    GAMMA FP16 & 0.8233 & 0.9433 & 0.9798 & 0.0131 \\
    GAMMA INT8 & 0.8232 & 0.9432 & 0.9798 & 0.0131 \\
    \midrule
    $\Delta$ INT8--FP32
      & $-0.0001$ & $-0.0001$ & $0.0000$ & $0.0000$ \\
    \bottomrule
  \end{tabular}
\end{table}





\bibliographystyle{IEEEtran}
\bibliography{references}

@article{geiger2013vision,
  title={Vision meets robotics: The kitti dataset},
  author={Geiger, Andreas and Lenz, Philip and Stiller, Christoph and Urtasun, Raquel},
  journal={The international journal of robotics research},
  volume={32},
  number={11},
  pages={1231--1237},
  year={2013},
  publisher={Sage Publications Sage UK: London, England}
}

@inproceedings{zhou2018voxelnet,
  title={Voxelnet: End-to-end learning for point cloud based 3d object detection},
  author={Zhou, Yin and Tuzel, Oncel},
  booktitle={Proceedings of the IEEE conference on computer vision and pattern recognition},
  pages={4490--4499},
  year={2018}
}

@article{yan2018second,
  title={Second: Sparsely embedded convolutional detection},
  author={Yan, Yan and Mao, Yuxing and Li, Bo},
  journal={Sensors},
  volume={18},
  number={10},
  pages={3337},
  year={2018},
  publisher={MDPI}
}

@inproceedings{caesar2020nuscenes,
  title={nuscenes: A multimodal dataset for autonomous driving},
  author={Caesar, Holger and Bankiti, Varun and Lang, Alex H and Vora, Sourabh and Liong, Venice Erin and Xu, Qiang and Krishnan, Anush and Pan, Yu and Baldan, Giancarlo and Beijbom, Oscar},
  booktitle={Proceedings of the IEEE/CVF conference on computer vision and pattern recognition},
  pages={11621--11631},
  year={2020}
}

@ARTICLE{apalffy2022,
  author={Palffy, Andras and Pool, Ewoud and Baratam, Srimannarayana and Kooij, Julian F. P. and Gavrila, Dariu M.},
  journal={IEEE Robotics and Automation Letters}, 
  title={Multi-Class Road User Detection With 3+1D Radar in the View-of-Delft Dataset}, 
  year={2022},
  volume={7},
  number={2},
  pages={4961-4968},
  doi={10.1109/LRA.2022.3147324}}

@inproceedings{xu2025mergeocc,
  title={MergeOcc: Bridge the Domain Gap between Different LiDARs for Robust Occupancy Prediction},
  author={Xu, Zikun and Xu, Shaobing},
  booktitle={Proceedings of the IEEE/CVF International Conference on Computer Vision},
  pages={26539--26548},
  year={2025}
}

@inproceedings{mescheder2019occupancy,
  title={Occupancy networks: Learning 3d reconstruction in function space},
  author={Mescheder, Lars and Oechsle, Michael and Niemeyer, Michael and Nowozin, Sebastian and Geiger, Andreas},
  booktitle={Proceedings of the IEEE/CVF conference on computer vision and pattern recognition},
  pages={4460--4470},
  year={2019}
}

@inproceedings{park2019deepsdf,
  title={Deepsdf: Learning continuous signed distance functions for shape representation},
  author={Park, Jeong Joon and Florence, Peter and Straub, Julian and Newcombe, Richard and Lovegrove, Steven},
  booktitle={Proceedings of the IEEE/CVF conference on computer vision and pattern recognition},
  pages={165--174},
  year={2019}
}

@inproceedings{biber2003normal,
  title={The normal distributions transform: A new approach to laser scan matching},
  author={Biber, Peter and Stra{\ss}er, Wolfgang},
  booktitle={Proceedings 2003 IEEE/RSJ International Conference on Intelligent Robots and Systems (IROS 2003)(Cat. No. 03CH37453)},
  volume={3},
  pages={2743--2748},
  year={2003},
  organization={IEEE}
}

@article{kerbl20233d,
  title={3d gaussian splatting for real-time radiance field rendering.},
  author={Kerbl, Bernhard and Kopanas, Georgios and Leimk{\"u}hler, Thomas and Drettakis, George and others},
  journal={ACM Trans. Graph.},
  volume={42},
  number={4},
  pages={139--1},
  year={2023}
}

@inproceedings{hess2025splatad,
  title={Splatad: Real-time lidar and camera rendering with 3d gaussian splatting for autonomous driving},
  author={Hess, Georg and Lindstr{\"o}m, Carl and Fatemi, Maryam and Petersson, Christoffer and Svensson, Lennart},
  booktitle={Proceedings of the Computer Vision and Pattern Recognition Conference},
  pages={11982--11992},
  year={2025}
}

@inproceedings{eckart2016accelerated,
  title={Accelerated generative models for 3d point cloud data},
  author={Eckart, Benjamin and Kim, Kihwan and Troccoli, Alejandro and Kelly, Alonzo and Kautz, Jan},
  booktitle={Proceedings of the IEEE Conference on Computer Vision and Pattern Recognition},
  pages={5497--5505},
  year={2016}
}

@inproceedings{boulch2023also,
  title={Also: Automotive lidar self-supervision by occupancy estimation},
  author={Boulch, Alexandre and Sautier, Corentin and Michele, Bj{\"o}rn and Puy, Gilles and Marlet, Renaud},
  booktitle={Proceedings of the IEEE/CVF Conference on Computer Vision and Pattern Recognition},
  pages={13455--13465},
  year={2023}
}

@inproceedings{qi2017pointnet,
  title={Pointnet: Deep learning on point sets for 3d classification and segmentation},
  author={Qi, Charles R and Su, Hao and Mo, Kaichun and Guibas, Leonidas J},
  booktitle={Proceedings of the IEEE conference on computer vision and pattern recognition},
  pages={652--660},
  year={2017}
}

@article{qi2017pointnet++,
  title={Pointnet++: Deep hierarchical feature learning on point sets in a metric space},
  author={Qi, Charles Ruizhongtai and Yi, Li and Su, Hao and Guibas, Leonidas J},
  journal={Advances in neural information processing systems},
  volume={30},
  year={2017}
}

@inproceedings{yang2018foldingnet,
  title={Foldingnet: Point cloud auto-encoder via deep grid deformation},
  author={Yang, Yaoqing and Feng, Chen and Shen, Yiru and Tian, Dong},
  booktitle={Proceedings of the IEEE conference on computer vision and pattern recognition},
  pages={206--215},
  year={2018}
}

@inproceedings{groueix2018papier,
  title={A papier-m{\^a}ch{\'e} approach to learning 3d surface generation},
  author={Groueix, Thibault and Fisher, Matthew and Kim, Vladimir G and Russell, Bryan C and Aubry, Mathieu},
  booktitle={Proceedings of the IEEE conference on computer vision and pattern recognition},
  pages={216--224},
  year={2018}
}

@article{pang2023masked,
  title={Masked autoencoders for 3d point cloud self-supervised learning},
  author={Pang, Yatian and Tay, Eng Hock Francis and Yuan, Li and Chen, Zhenghua},
  journal={World Scientific Annual Review of Artificial Intelligence},
  volume={1},
  pages={2440001},
  year={2023},
  publisher={World Scientific}
}

@inproceedings{yu2022point,
  title={Point-bert: Pre-training 3d point cloud transformers with masked point modeling},
  author={Yu, Xumin and Tang, Lulu and Rao, Yongming and Huang, Tiejun and Zhou, Jie and Lu, Jiwen},
  booktitle={Proceedings of the IEEE/CVF conference on computer vision and pattern recognition},
  pages={19313--19322},
  year={2022}
}

@article{min2022voxel,
  title={Voxel-mae: Masked autoencoders for pre-training large-scale point clouds},
  author={Min, Chen and Zhao, Dawei and Xiao, Liang and Nie, Yiming and Dai, Bin},
  journal={arXiv preprint arXiv:2206.09900},
  volume={2},
  number={3},
  year={2022}
}

@inproceedings{song2017semantic,
  title={Semantic scene completion from a single depth image},
  author={Song, Shuran and Yu, Fisher and Zeng, Andy and Chang, Angel X and Savva, Manolis and Funkhouser, Thomas},
  booktitle={Proceedings of the IEEE conference on computer vision and pattern recognition},
  pages={1746--1754},
  year={2017}
}

@inproceedings{roldao2020lmscnet,
  title={Lmscnet: Lightweight multiscale 3d semantic completion},
  author={Roldao, Luis and De Charette, Raoul and Verroust-Blondet, Anne},
  booktitle={2020 International Conference on 3D Vision (3DV)},
  pages={111--119},
  year={2020},
  organization={IEEE}
}

@article{chen2024lidar,
  title={Lidar-gs: Real-time lidar re-simulation using gaussian splatting},
  author={Chen, Qifeng and Yang, Sheng and Du, Sicong and Tang, Tao and Xie, Rengan and Chen, Peng and Huo, Yuchi},
  journal={arXiv preprint arXiv:2410.05111},
  year={2024}
}

@inproceedings{lin2017focal,
  title={Focal loss for dense object detection},
  author={Lin, Tsung-Yi and Goyal, Priya and Girshick, Ross and He, Kaiming and Doll{\'a}r, Piotr},
  booktitle={Proceedings of the IEEE international conference on computer vision},
  pages={2980--2988},
  year={2017}
}

@inproceedings{milletari2016v,
  title={V-net: Fully convolutional neural networks for volumetric medical image segmentation},
  author={Milletari, Fausto and Navab, Nassir and Ahmadi, Seyed-Ahmad},
  booktitle={2016 fourth international conference on 3D vision (3DV)},
  pages={565--571},
  year={2016},
  organization={Ieee}
}

@article{loshchilov2017decoupled,
  title={Decoupled weight decay regularization},
  author={Loshchilov, Ilya and Hutter, Frank},
  journal={arXiv preprint arXiv:1711.05101},
  year={2017}
}

@article{loshchilov2016sgdr,
  title={Sgdr: Stochastic gradient descent with warm restarts},
  author={Loshchilov, Ilya and Hutter, Frank},
  journal={arXiv preprint arXiv:1608.03983},
  year={2016}
}

@inproceedings{sun2020scalability,
  title={Scalability in perception for autonomous driving: Waymo open dataset},
  author={Sun, Pei and Kretzschmar, Henrik and Dotiwalla, Xerxes and Chouard, Aurelien and Patnaik, Vijaysai and Tsui, Paul and Guo, James and Zhou, Yin and Chai, Yuning and Caine, Benjamin and others},
  booktitle={Proceedings of the IEEE/CVF conference on computer vision and pattern recognition},
  pages={2446--2454},
  year={2020}
}

@article{choi2023msc,
  title={Msc-rad4r: Ros-based automotive dataset with 4d radar},
  author={Choi, Minseong and Yang, Seunghoon and Han, Seungho and Lee, Yeongseok and Lee, Minyoung and Choi, Keun Ha and Kim, Kyung-Soo},
  journal={IEEE Robotics and Automation Letters},
  volume={8},
  number={11},
  pages={7194--7201},
  year={2023},
  publisher={IEEE}
}

@inproceedings{zhong2023shine,
  title={Shine-mapping: Large-scale 3d mapping using sparse hierarchical implicit neural representations},
  author={Zhong, Xingguang and Pan, Yue and Behley, Jens and Stachniss, Cyrill},
  booktitle={2023 IEEE International Conference on Robotics and Automation (ICRA)},
  pages={8371--8377},
  year={2023},
  organization={IEEE}
}

@article{hornung2013octomap,
  title={OctoMap: An efficient probabilistic 3D mapping framework based on octrees},
  author={Hornung, Armin and Wurm, Kai M and Bennewitz, Maren and Stachniss, Cyrill and Burgard, Wolfram},
  journal={Autonomous robots},
  volume={34},
  number={3},
  pages={189--206},
  year={2013},
  publisher={Springer}
}

\end{document}